\documentclass{article}

\usepackage{arxiv}

\usepackage[utf8]{inputenc} 
\usepackage[T1]{fontenc}    
\usepackage{hyperref}       
\usepackage{url}            
\usepackage{booktabs}       
\usepackage{amsfonts}       
\usepackage{nicefrac}       
\usepackage{microtype}      
\usepackage{lipsum}		
\usepackage{graphicx}
\usepackage{natbib}
\usepackage{doi}
\usepackage{amsmath}
\usepackage{multirow}

\title{Uncertainty-Guided Cross Attention Ensemble Mean Teacher for Semi-supervised Medical Image Segmentation}

\date{} 					
\author{ \href{}{\hspace{1mm}Meghana Karri}\\
	Machine and Hybrid Intelligence Lab\\
	Northwestern University\\
	Chicago, USA \\
	\texttt{meghana.karri@northwestern.edu} \\
	\And
	\href{}{\hspace{1mm}Amit Soni Arya} \\
	School of Computer Science Engineering and Technology\\
	Bennett University\\
	India \\
	\texttt{amitsoniuoh@gmail.com} \\
    \And
	\href{}{\hspace{1mm}Koushik Biswas} \\
	Machine and Hybrid Intelligence Lab\\
	Northwestern University\\
	Chicago, USA \\
	\texttt{koushik.biswas@northwestern.edu} \\
    \And
	\href{}{\hspace{1mm}Nicolo Gennaro} \\
	Northwestern University\\
	Chicago, USA \\
	\texttt{nicolo.gennaro@northwestern.edu} \\
    \And
	\href{}{\hspace{1mm}Vedat Cicek, Gorkem Durak, Yury Velichko, Ulas Bagci} \\
	Machine and Hybrid Intelligence Lab\\
	Northwestern University\\
	Chicago, USA \\
	\texttt{\{vedat.cicek, gorkem.durak, y-velichko, ulas.bagci\}@northwestern.edu} \\
}

\renewcommand{\shorttitle}{Semi supervised medical image segmentation}

\hypersetup{
pdftitle={Uncertainty-Guided Cross Attention Ensemble Mean Teacher for Semi-supervised Medical Image Segmentation},
pdfsubject={q-bio.NC, q-bio.QM},
pdfauthor={David S.~Hippocampus, Elias D.~Striatum},
pdfkeywords={First keyword, Second keyword, More},
}

\begin{document}
\maketitle

\begin{abstract}
	This work proposes a novel framework, Uncertainty-Guided Cross Attention Ensemble Mean Teacher (UG-CEMT), for achieving state-of-the-art performance in semi-supervised medical image segmentation. UG-CEMT leverages the strengths of co-training and knowledge distillation by combining a Cross-attention Ensemble Mean Teacher framework (CEMT) inspired by Vision Transformers (ViT) with uncertainty-guided consistency regularization and Sharpness-Aware Minimization emphasizing uncertainty. UG-CEMT improves semi-supervised performance while maintaining a consistent network architecture and task setting by fostering high disparity between sub-networks. Experiments demonstrate significant advantages over existing methods like Mean Teacher and Cross-pseudo Supervision in terms of disparity, domain generalization, and medical image segmentation performance. UG-CEMT achieves state-of-the-art results on multi-center prostate MRI and cardiac MRI datasets, where object segmentation is particularly challenging. Our results show that using only 10\% labeled data, UG-CEMT approaches the performance of fully supervised methods, demonstrating its effectiveness in exploiting unlabeled data for robust medical image segmentation. The code is publicly available at \url{https://github.com/Meghnak13/UG-CEMT}
\end{abstract}


\begin{figure}
    \centering
    \includegraphics[width=0.7\linewidth]{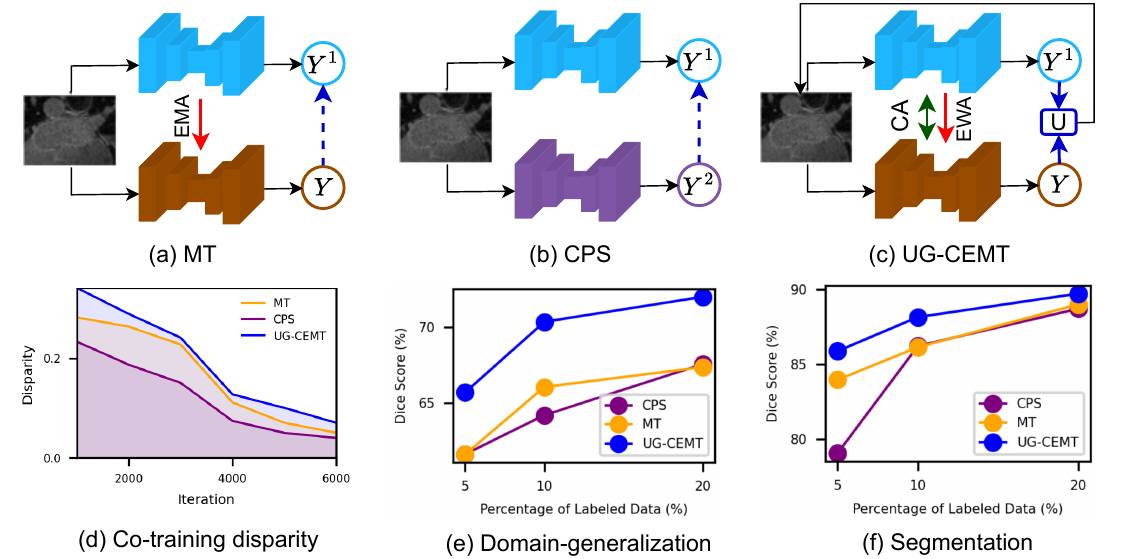}
    \caption{Comparison of architectures and their performance for SSL segmentation tasks: (a) Mean Teacher (MT), (b) Cross-Pseudo supervision (CPS), (c) UG-CEMT framework (proposed), (d) disparity between co-training sub-networks w.r.t Jaccard metric, (e) domain generalization effectiveness for multi-site prostate dataset, and (f) segmentation performance on single-site LA dataset. }
    \label{fig:comp}
\end{figure}
\section{Introduction}
\label{sec:intro}
Medical image segmentation, essential for accurate diagnosis and treatment, has significantly benefited from recent advancements in deep learning, achieving higher accuracy and automation~\cite{chen2018encoder, ronneberger2015u}. However, training these models requires a large amount of labeled data, a laborious process requiring expert knowledge and prone to human error and bias. As a result, labeled data is often limited to well-studied structures like major organs or common cancers, while unlabeled data is readily available and less expensive to acquire. This has motivated the exploration of semi-supervised learning (SSL) techniques~\cite{van2020survey}. SSL leverages both abundant unlabeled data and limited labeled data, enhancing model generalizability, reducing annotation burden, and unlocking the power of unlabeled data for various medical imaging tasks, including anomaly detection, disease segmentation, and classification~\cite{berthelot2019mixmatch, cascante2021curriculum, luo2018smooth, yang2021instance}. However, despite its potential, SSL for medical imaging faces several key challenges: data quality and mislabels, limited control over large-scale unlabeled data, bias due to data selection, explainability issues, and domain generalization problems. Following these challenges, various SSL methods have been proposed, with consistency regularization~\cite{tarvainen2017mean, xie2020unsupervised} and pseudo-labeling~\cite{lee2013pseudo, yao2022enhancing, shen2023co} are the two mainstream SSL approaches used for medical image segmentation. 
In this study, we propose an alternative method combining the complementary strengths of these two primary approaches via cross-supervision and knowledge distillation (i.e., Teacher-Student sub-networks). Specifically, we propose an \textit{Uncertainty-Guided Cross Attention Ensemble Mean Teacher (UG-CEMT)} framework, which excels at maintaining higher disparity between co-training segmentation sub-networks by leveraging high-confidence predictions. This, in turn, enhances semi-supervised segmentation performance while using a consistent backbone network and task settings. 
Unlike traditional approaches that rely heavily on pseudo generation, our UG-CEMT framework prioritizes consistency regularization, which enforces the model to maintain consistent predictions across different perturbations of the same input data. This focus on consistency over prediction generation helps to reduce the propagation of errors that can occur when low-confidence predictions are used during training. For instance, while standard mean teacher frameworks like UA-MT~\cite{yu2019uncertainty} primarily utilize a static approach to uncertainty estimation, UG-CEMT dynamically adapts its learning process through a novel cross-attention mechanism and two-step training process, ensuring robust feature alignment and improving generalization. Moreover, unlike approaches such as MCF-Net~\cite{wang2023mcf}, which may employ pseudo-labels without considering uncertainty, UG-CEMT leverages uncertainty-guided consistency using the Monte Carlo dropout (MC Dropout) to prioritize high-confidence predictions, significantly enhancing segmentation performance. Additionally, Sharpness-Aware Minimization (SAM) regularization further enhances the model’s generalization by smoothing the loss landscape and incorporating uncertainty to improve robustness. 
Figure~\ref{fig:comp} illustrates high-level comparisons of architectures and their performance for SSL segmentation tasks: mean teacher (MT)~\cite{tarvainen2017mean}, cross-pseudo supervision (CPS)~\cite{chen2021semi}, and our proposed method UG-CEMT. Disparity between co-training networks, effectiveness of these methods on domain generalization, and medical image segmentation are also illustrated. In the experiments, we show detailed results supporting our observations and findings.%

Summary of our contributions are as follows:
\begin{itemize} \item We identify and address critical limitations in existing co-training-based semi-supervised segmentation approaches, specifically the inadequate disparity among sub-networks and reliance on low-confidence predictions. UG-CEMT introduces a novel cross-attention mechanism that dynamically enhances disparity and an uncertainty-guided consistency strategy that prioritizes high-confidence predictions, collectively leading to superior segmentation performance. \item Our framework’s two-step training process, driven by uncertainty estimation, not only improves the initial training phase but also guides the model in refining its predictions in the subsequent phase. This process, combined with SAM regularization, ensures that the model remains robust across different domains and data variations. \item Comprehensive experiments across various public medical image segmentation datasets, including challenging 3D scenarios, validate the novel integration of cross-attention and uncertainty-guided regularization in UG-CEMT. Our results demonstrate the clear superiority of UG-CEMT over existing state-of-the-art approaches, particularly in maintaining network disparity and enhancing domain generalization. \end{itemize}

\textbf{Clinical significance of the problem:} 
This study addresses two critical tasks: cardiac MRI and prostate MRI analysis. Cardiac MRI is vital for diagnosing and monitoring cardiovascular diseases, the leading cause of mortality worldwide. Accurate left atrium (LA) segmentation is key for evaluating cardiac conditions like atrial fibrillation~\cite{aliasghar}, but is challenging due to size, shape variations, and inconsistent image quality. Prostate cancer, the most diagnosed cancer in men and the fifth leading cause of cancer deaths globally~\cite{turkbey}, also relies on MRI for detection and staging. Segmentation is difficult due to the similar density of surrounding tissues and variability in image quality across centers. Our work uses multi-center prostate MRI data to assess the generalization of the proposed SSL system.

\section{Related Work}
The objective of SSL is to enhance the effectiveness of supervised learning (SL) by utilizing unlabeled data in conjunction with labeled data~\cite{van2020survey}. A prevalent method in SSL is the inclusion of a regularization factor in the SL objective function, which enables the model to capitalize on unlabeled data. In this regard, SSL algorithms can be broadly classified into two primary approaches: \textit{consistency regularization} \cite{tarvainen2017mean} and \textit{pseudo-labeling} \cite{lee2013pseudo}. Pseudo-labeling attempts to generate pseudo-labels for unlabeled data, mimicking ground truth labels used in supervised training. Consistency regularization, on the other hand, enforces the model's predictions to remain consistent across different input variations. Both of these techniques have been successfully applied to SSL for image classification, achieving exceptional results \cite{sohn2020fixmatch, zhang2021flexmatch}.

\textbf{Challenges of SSL:} Many SSL methods leverage labeled data supervision not just for initialization or training convergence, but as a crucial element to explicitly guide knowledge extraction from unlabeled data~\cite{miyato2018virtual}. This is particularly relevant in the context of predominantly labeled clinical datasets, where foreground features such as appearance, shape, and texture are often consistent across diverse samples. By bridging the gap between labeled and unlabeled data within the entire training set, SSL has the potential to effectively transfer prior knowledge from labeled examples to unlabeled data,  overcoming the performance limitations often encountered in SSL approaches.

\textbf{Co-teaching methods:} Recent approaches in SSL utilize mutual learning or co-teaching paradigms to achieve promising results~\cite{yu2019does,chen2021semi}. These methods combine consistency regularization with entropy minimization. They employ two models that are trained simultaneously, with each model predicting the output of its counterpart. This approach has shown significant improvement in segmentation performance for semi-supervised medical image segmentation, as demonstrated by MC-Net~\cite{seibold2022reference}.

\textbf{Consistency regularization:} Consistency regularization, exemplified by MT~\cite{tarvainen2017mean}, enforces consistent predictions across perturbed inputs for a student-teacher network pair via gradient descent and an exponential weighted average (EWA), respectively. Subsequent methods have built on this idea. For example, \cite{luo2018smooth} introduced a graph-based method to ensure adjacent points remain consistent under perturbations. Miyato et al. \cite{miyato2018virtual} incorporated adversarial perturbations into consistency learning, leading to interpolation-consistent training (ICT) \cite{verma2022interpolation} to avoid potential generalization issues. Wang et al. \cite{wang2021tripled} combined multitask learning with MT, using triple uncertainty to guide the student model. Huang et al. \cite{huang2022semi} proposed a method for neuron segmentation based on pixel-level prediction consistency. However, these methods often overlook the interactions between sub-networks and may struggle to address inherent network biases. \textit{Our proposed UG-CEMT framework addresses these limitations by employing a cross-attention mechanism to enhance feature alignment and information exchange between student and teacher models. }

\textbf{Problem of low-confidence pseudo labels:} While consistency regularization is an effective SSL method, current co-training models using consistency regularization often converge rapidly to a consensus, leading to low-confidence pseudo labels from perturbed input data during training. This premature convergence results in the model degenerating into self-training. Maintaining disparity among sub-networks is essential for effective co-training, as it ensures that the information provided by each sub-network remains complementary. Furthermore, the effectiveness of these models is substantially influenced by the quality of pseudo labels, which should exhibit low uncertainty. 

\textbf{Uncertainty-guided semi-supervised learning:} 
Exploiting model uncertainty (epistemic) for consistency regularization offers promise in SSL. However, accurate estimation and effective utilization remain challenges. Common methods rely on MC Dropout~\cite{gal2016dropout} or prediction variance~\cite{zhang2021flexmatch}. Prior works leverage uncertainty for loss reweighting~\cite{yu2019uncertainty} or contrastive sample selection~\cite{wang2022semi}. These methods often require a fixed threshold for filtering low-confidence pseudo-labels, a process hampered by the difficulty of setting an appropriate value. Our UG-CEMT framework addresses this by estimating uncertainty via MC Dropout and entropy calculation. Instead of filtering pseudo-labels, we integrate this uncertainty into consistency regularization, enabling the model to learn robustly from both labeled and unlabeled data based on dynamic confidence levels. This eliminates the need for fixed thresholds, ultimately improving segmentation performance.

\section{Methods}
\begin{figure*}[!t]
\begin{minipage}[t]{.55\linewidth}
        \centering
         \includegraphics[height=14cm, width=9.6cm, keepaspectratio]{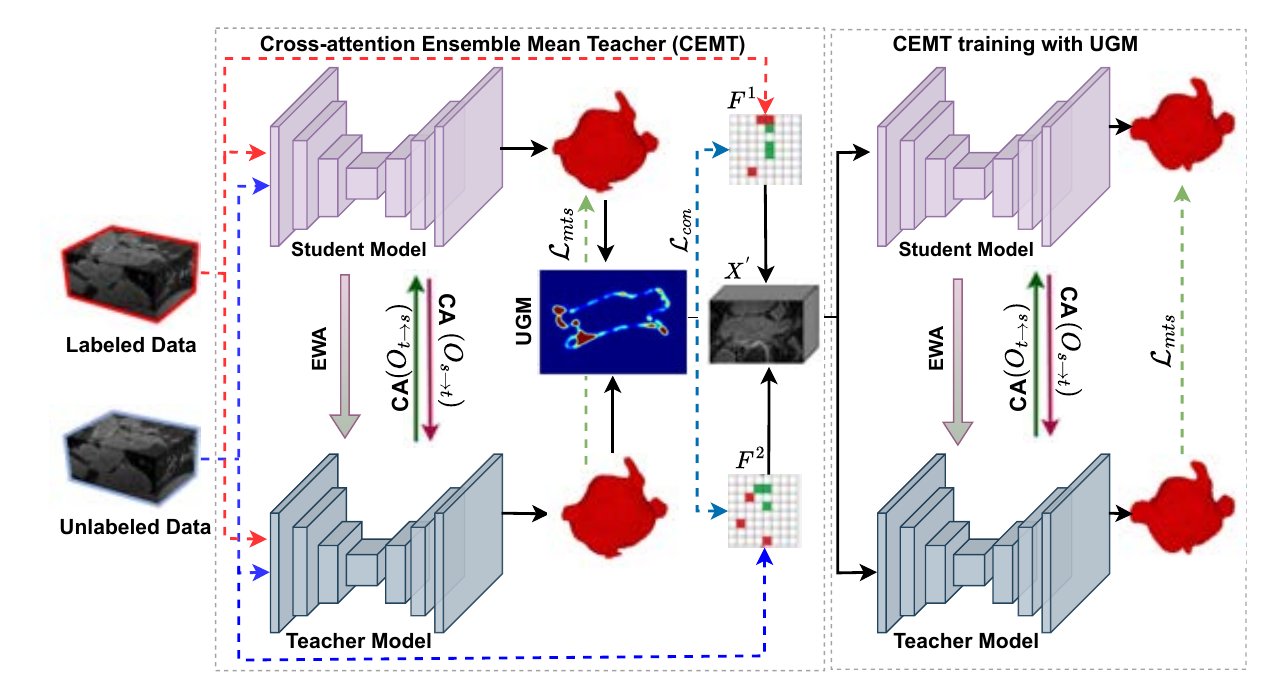}
        
        \caption{The proposed UG-CEMT architecture. UG-CEMT creates new samples $X^{'}$ from input data using UGM. Cross-Attention (CA) is applied between the student and teacher model, where $O_{(s\rightarrow t)}$ and $O_{(t\rightarrow s)}$ represent outputs of attention mechanism from student to teacher, and teacher to student respectively.}
        \label{fig:enter-label}
         \end{minipage}
         \hfill
   \begin{minipage}[t]{.43\linewidth}
        \centering
    
         \includegraphics[width=\linewidth]{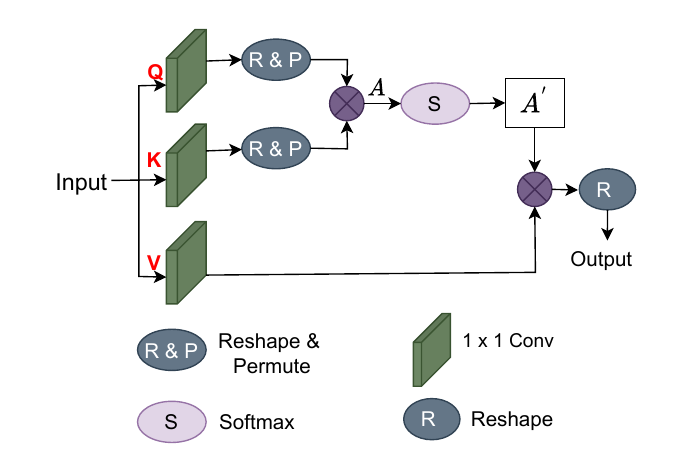}
        
        \caption{Overview of the proposed cross-attention (CA) mechanism (inspired by ViT).}
        \label{fig:ca}
   \end{minipage}
\end{figure*}

Our proposed UG-CEMT framework addresses critical challenges in existing co-training-based semi-supervised segmentation methods by enhancing model disparity among sub-networks, leveraging a cross-attention mechanism, an uncertainty-guided consistency regularization process, and SAM regularization. These components collectively ensure that our model effectively utilizes both labeled and unlabeled data for superior segmentation performance. As illustrated in Figure~\ref{fig:enter-label}, the proposed model includes a two-step training phase. In the first step, we train the CEMT using both labeled and unlabeled data to generate uncertainty-guided maps. In the CEMT framework, the uncertainty-guided map ensures low uncertainty for high-confidence predictions. In the second step, we retrain the CEMT model using these uncertainty-guided maps to further enhance segmentation performance.

\subsection{Cross-attention (CA) mechanism}
As shown in Figure \ref{fig:ca}, the cross-attention mechanism in our UG-CEMT framework is designed to robustly align features and facilitate information exchange between the student and teacher models, thereby enhancing overall segmentation performance. For the student model’s feature map $X_{s}$ and the teacher model’s feature map $X_{t}$, we first project these features into queries $(Q)$, keys $(K)$, and values $(V)$ via learnable weight matrices $W_{Q}, W_{K}, W_{V}$ as:$  \quad \quad Q_{s}=X_{s}W_{Q}^{s}, \quad  K_{t}=X_{t}W_{K}^{t}, \quad  V_{t}=X_{t}W_{V}^{t},
  Q_{t}=X_{t}W_{Q}^{t}, \quad  K_{s}=X_{s}W_{K}^{s}, \quad  V_{s}=X_{s}W_{V}^{s}.$
The cross-attention from the student model to the teacher model is then computed by taking the dot product of the student queries $Q_{s}$
and the teacher keys $K_{t}$, scaling by the square root of the key dimension 
$\sqrt{d_{k}}$, and applying a softmax function to obtain the attention weights. These weights are then used to compute a weighted sum of the teacher values $V_{t}$:
\begin{equation}
    CA_{s\rightarrow t}(Q_{s}, K_{t}, V_{t}) = Softmax \left(\frac{Q_{s}K_{t}^{T}}{\sqrt{d_{k}}}\right).
\end{equation}
Similarly, the cross-attention (CA) from the teacher model to the student model is computed as:
\begin{equation}
    CA_{t\rightarrow s}(Q_{t}, K_{s}, V_{s}) = Softmax \left(\frac{Q_{t}K_{s}^{T}}{\sqrt{d_{k}}}\right).
\end{equation}
These attention weights are used to compute a weighted sum of the values:
\begin{equation}
    O_{s\rightarrow t}=CA_{s\rightarrow t}(Q_{s}, K_{t}, V_{t})V_{t},
\end{equation}
\begin{equation}
    O_{t\rightarrow s}=CA_{s\rightarrow t}(Q_{t}, K_{s}, V_{s})V_{s}.
\end{equation}
The output of the attention mechanism is combined with the original feature maps using another learnable parameter $\gamma$:$X_{s}^{'}=\gamma \dot O_{s\rightarrow t} + X_{s}, \text{ and }  X_{t}^{'}=\gamma \dot O_{t\rightarrow s} + X_{t}.$
The updated features $X_{s}^{'}$ and $X_{t}^{'}$ are then used in subsequent layers of the student and teacher models, respectively.

\subsection{Semi-supervised segmentation}
Assuming a training set with $N$ labeled and $M$ unlabeled examples. Let the labeled dataset be denoted as $\mathcal{D}_{L}=\left\{(a_{i},b_{i})\right\}_{i=1}^{M}$, and the unlabeled dataset be represented as $\mathcal{D}_{U}=\left\{a_{i}\right\}_{i=M+1}^{M+N}$. Here $a_{i}\in \mathbb{R}^{H\times W\times D}$ refers to the input image (volume), and $b_{i}\in \{0,1\}^{H\times W\times D}$ indicates the corresponding ground-truth annotation (i.e., segmentation maps). The primary objective of our semi-supervised segmentation framework is to minimize a combined loss function that integrates both supervised and unsupervised components: 

\begin{equation}
    min_{\theta } \bigg( \sum_{i=1}^{M} \mathcal{L}_{s}\left[ f\left( a_{i}; \theta \right), b_{i} \right ] +  \lambda \sum_{j=M+1}^{M+N} \mathcal{L}_{c} \left[ f( a_{j}; \theta ,\zeta ),f( a_{j}; \theta ,\zeta^{'} )\right ] \bigg )
\end{equation}
Where $\mathcal{L}_{s}$ represents the supervised loss, such as cross-entropy loss, which evaluates the accuracy of the network's output against the labeled data. $\mathcal{L}_{c}$ indicates the unsupervised consistency loss, which quantifies the agreement between the predictions of the student and teacher sub-networks for the same input $a_{j}$ under different perturbations. The segmentation network is denoted by $f(\cdot)$, and $(\theta, \zeta)$, $(\theta, \zeta^{'})$ pair represent the model and perturbations (e.g., input noise and dropout) parameters applied to the teacher and student sub-networks, respectively. The term $\lambda$ indicates a ramp-up coefficient that balances the supervised and unsupervised loss components.

\subsection{Uncertainty-guided consistency regularization}
The Uncertainty-Guided Consistency Regularization framework leverages the concept of uncertainty to enhance the robustness and reliability of the semi-supervised learning process. This framework estimates model uncertainty and integrates it with the mean teacher model to improve the training of the student model.

\textbf{1) Uncertainty estimation:} While various techniques exist for estimating uncertainty in deep learning models, we opt for MC Dropout due to its several advantages. MC Dropout offers a well-established and computationally efficient approach, allowing us to leverage the existing network architecture without introducing significant modifications. Additionally, MC Dropout provides inherent interpretability by directly reflecting the model's confidence in its predictions. This interpretability is crucial in our setting, where understanding and prioritizing high-confidence predictions for uncertainty-guided consistency regularization is essential~\cite{isler}. Herein, we employ MC dropout during training by performing multiple stochastic forward passes with dropout to generate a set of predictions for each input. The uncertainty is then quantified using the entropy of these predictions. For an input $x$, we obtain $N$ stochastic predictions $\{\hat{y}_{1}, \hat{y}_{2}, \cdots , \hat{y}_{N}\}$ by applying dropout: $
\hat{y}_{mean}=\frac{1}{N}\sum_{i=1}^{N}\hat{y}_{i}$
The entropy of the mean prediction is calculated to quantify the uncertainty:
\begin{equation}
Entropy(\hat{y}_{mean}) = -{\sum_{c}}\hat{y}^{c}_{mean}log (\hat{y}^{c}_{mean}),
\end{equation}
where $c$ denotes the class index. Higher entropy indicates greater uncertainty in the prediction.

\textbf{2) Ensemble mean teacher model:} Recent research has demonstrated that ensemble mean teacher predictions from different training stages can enhance prediction quality. By leveraging these ensembled predictions as the teacher's predictions, we can achieve improved segmentation results. Consequently, the teacher model's weights $\theta_{t}$ are updated using an exponential weighted average (EWA) of the student model's weights $\theta$. This EWA approach ensures that the teacher model captures information from different stages of the training process. The update rule for the teacher model's weights at training step $t$ is given by: $\theta _{t}=\beta \theta _{t-1}+(1-\beta )\theta _{t}$, where $\beta$ is the EWA decay rate controlling the update pace. 

\textbf{3) Consistency regularization with uncertainty guidance:} Consistency regularization encourages the student model to produce similar predictions under different perturbations. In the UG-CEMT framework, this regularization is guided by the estimated uncertainty. Specifically, we focus on areas with low uncertainty to enforce consistency, as these areas are more likely to provide reliable supervision signals. The consistency loss $\mathcal{L}_{cons}$ is defined as:
\begin{equation}
    \mathcal{L}_{cons} = \mathbb{E}_{x\sim u}\left [ U(x)\left\| f_{s}(x)-f_{t}(x^{'})\right\|^{2}_{2} \right ],
\end{equation}
where $U$ represents the distribution of unlabeled data, $x^{'}$ is a perturbed version of $x$, and $U(x)$ is the uncertainty weight based on the entropy: $ U(x)=exp(-Entropy(\hat{y}_{mean}))$.
This weighting ensures that regions with lower uncertainty have a higher impact on the consistency loss, encouraging the model to learn from more reliable predictions.

\subsection{Overall training objective}
We utilize V-Net~\cite{milletari2016v} as our backbone network and modify it by removing the short residual connections in each convolution block. Our training objective combines dice and cross-entropy loss. To enable V-Net for uncertainty estimation, dropout layers with a dropout rate of 0.5 are added after both the R-Stage 1 and L-stage 5 layers. Dropout is activated during training and uncertainty estimation but deactivated during testing. We set the EWA decay parameter $\alpha$ to 0.99, consistent with previous research. A time-dependent Gaussian warming-up function $\phi(t)=0.1\times e^{(-5(1-\frac{t}{t_{max}})^{2})}$ is used to balance the supervised and unsupervised consistency losses, where $t$ is the current training step and $t_{max}$ is the maximum training step~\cite{li2020shape, wu2021semi}. This approach ensures that the supervised loss term dominates at the start, preventing the network from becoming stuck in a degenerate solution with no meaningful target predictions for unlabeled data. For uncertainty estimation, we set $T=8$ to balance the quality of uncertainty estimation with training efficiency. As training progresses, this method allows the student model to gradually learn from increasingly uncertain cases. To further enhance generalization and robustness, we incorporate the SAM optimizer~\cite{wu2024cr}. SAM optimizes the model weights by not only minimizing the training loss but also ensuring the loss landscape is smooth (by acting as a regularizer). This is achieved by performing a gradient ascent step followed by a gradient descent step, effectively flattening the minima in the loss surface. The SAM optimizer requires careful tuning of its parameters, such as the radius of the neighborhood $\rho$, which we empirically set to achieve a balance between performance and stability. For our implementation, we use a SAM optimizer with $\rho=0.5$.

\section{Experiments and Results}
\textbf{Datasets:} We evaluate our model on two challenging medical image segmentation datasets. First, publicly available \textit{LA-Dataset }(3D Left Atrium Segmentation challenge)~\cite{xiong2021global} comprises 100 3D (volumetric) gadolinium-enhanced MRI scans (GE-MRIs) with corresponding left atrium (LA) labels for training and validation. All images were cropped to center on the heart location and normalized. We split the data into 80 images for training and 20 for testing by following challenge guidelines. Second, \textit{Multi-Site Prostate MRI Segmentation} dataset~\cite{liu2020ms} is designed to assess model robustness by containing T2-weighted MRI scans from six different medical centers (multi-site) with distinct data distributions. This dataset includes data from NCI-ISBI 2013, I2CVB, and PROMISE12 datasets, further separated by acquisition site. In total, there are 116 image volumes: 30 cases from each of RUNMC and BMC sites, 30 from NCI-ISBI2013's HCRUDB19 site, and 13, 12, and 12 cases from UCL, BIDME, and HK sites within PROMISE12, respectively. We split the data into 92 images for training and 24 for testing by following the challenge guidelines. \textbf{Evaluation metrics:} We use four performance measures to assess our model's effectiveness, including edge sensitive metrics: Average Surface Distance (ASD) and 95\% Hausdorff Distance (95HD), and regional sensitive metrics: Jaccard similarity coefficient (Jaccard) and Dice similarity coefficient (Dice).

\begin{table}[h!]
\centering
\caption{Comparison of the proposed UG-CEMT with other state-of-the-art SSL methods on LA dataset for 6000 iterations.} \label{sota-la}

\resizebox{10cm}{!}{ 
\begin{tabular}{l ll llll }
\hline
\multicolumn{1}{c}{\multirow{2}{*}{Method}} & \multicolumn{2}{c}{(\%) of images used} & \multicolumn{4}{c}{Metrics} \\ \cline{2-7} 
\multicolumn{1}{c}{} & \multicolumn{1}{l}{Labeled} & Unlabeled & \multicolumn{1}{l}{Dice $\uparrow$} & \multicolumn{1}{l}{Jaccard $\uparrow$} & \multicolumn{1}{l}{95HD $\downarrow$} & ASD $\downarrow$\\ \hline
B-VNet & \multicolumn{1}{l}{80(100\%)} & 0 & \multicolumn{1}{l}{\textbf{91.20}} & \multicolumn{1}{l}{\textbf{83.05}} & \multicolumn{1}{l}{\textbf{4.56}} & 1.95 \\ 
V-VNet & \multicolumn{1}{l}{80(100\%)} & 0 & \multicolumn{1}{l}{90.96} & \multicolumn{1}{l}{82.89} & \multicolumn{1}{l}{5.00} & \textbf{1.72} \\ 
B-VNet & \multicolumn{1}{l}{16(20\%)} & 0 & \multicolumn{1}{l}{84.26} & \multicolumn{1}{l}{73.54} & \multicolumn{1}{l}{18.12} & 4.95 \\ 
V-VNet & \multicolumn{1}{l}{16(20\%)} & 0 & \multicolumn{1}{l}{83.11} & \multicolumn{1}{l}{72.47} & \multicolumn{1}{l}{14.77} & 3.82 \\ \hline
UA-MT & \multicolumn{1}{l}{4(5\%)} & 76 & \multicolumn{1}{l}{78.23} & \multicolumn{1}{l}{65.03} & \multicolumn{1}{l}{22.17} & 8.63 \\ 
DTC & \multicolumn{1}{l}{4(5\%)} & 76 & \multicolumn{1}{l}{80.16} & \multicolumn{1}{l}{67.88} & \multicolumn{1}{l}{21.45} & 7.18 \\ 
CPS & \multicolumn{1}{l}{4(5\%)} & 76 & \multicolumn{1}{l}{79.07} & \multicolumn{1}{l}{68.26} & \multicolumn{1}{l}{16.23} & 6.89 \\ 
SASSNet & \multicolumn{1}{l}{4(5\%)} & 76 & \multicolumn{1}{l}{80.21} & \multicolumn{1}{l}{67.01} & \multicolumn{1}{l}{21.64} & 7.20 \\ 
MC-Net & \multicolumn{1}{l}{4(5\%)} & 76 & \multicolumn{1}{l}{80.92} & \multicolumn{1}{l}{68.25} & \multicolumn{1}{l}{17.25} & 3.43 \\ 
MT & \multicolumn{1}{l}{4(5\%)} & 76 & \multicolumn{1}{l}{83.97} & \multicolumn{1}{l}{72.67} & \multicolumn{1}{l}{15.56} & 5.03 \\ 
CEMT(Ours) & \multicolumn{1}{l}{4(5\%)} & 76 & \multicolumn{1}{l}{85.23} & \multicolumn{1}{l}{75.16} & \multicolumn{1}{l}{5.12} & 1.32 \\ 
UG-CEMT(Ours) & \multicolumn{1}{l}{4(5\%)} & 76 & \multicolumn{1}{l}{\textbf{85.89}} & \multicolumn{1}{l}{\textbf{76.23}} & \multicolumn{1}{l}{\textbf{3.39}} & \textbf{0.69} \\ \hline
UA-MT & \multicolumn{1}{l}{8(10\%)} & 72 & \multicolumn{1}{l}{85.81} & \multicolumn{1}{l}{75.41} & \multicolumn{1}{l}{18.25} & 5.04 \\ 
DTC & \multicolumn{1}{l}{8(10\%)} & 72 & \multicolumn{1}{l}{84.55} & \multicolumn{1}{l}{73.91} & \multicolumn{1}{l}{13.80} & 3.69 \\ 
CPS & \multicolumn{1}{l}{8(10\%)} & 72 & \multicolumn{1}{l}{86.23} & \multicolumn{1}{l}{76.22} & \multicolumn{1}{l}{11.68} & 3.65 \\ 
SASSNet & \multicolumn{1}{l}{8(10\%)} & 72 & \multicolumn{1}{l}{85.71} & \multicolumn{1}{l}{75.13} & \multicolumn{1}{l}{14.60} & 4.00 \\ 
MC-Net & \multicolumn{1}{l}{8(10\%)} & 72 & \multicolumn{1}{l}{85.13} & \multicolumn{1}{l}{77.49} & \multicolumn{1}{l}{10.35} & 1.85 \\ 
MT & \multicolumn{1}{l}{8(10\%)} & 72 & \multicolumn{1}{l}{86.15} & \multicolumn{1}{l}{76.16} & \multicolumn{1}{l}{11.37} & 3.60 \\ 
CEMT(Ours) & \multicolumn{1}{l}{8(10\%)} & 72 & \multicolumn{1}{l}{87.03} & \multicolumn{1}{l}{78.26} & \multicolumn{1}{l}{3.39} & 0.67 \\ 
UG-CEMT(Ours) & \multicolumn{1}{l}{8(10\%)} & 72 & \multicolumn{1}{l}{\textbf{88.16}} & \multicolumn{1}{l}{\textbf{79.83}} & \multicolumn{1}{l}{\textbf{3.08}} & \textbf{0.51} \\ \hline
UA-MT & \multicolumn{1}{l}{16(20\%)} & 64 & \multicolumn{1}{l}{88.13} & \multicolumn{1}{l}{78.04} & \multicolumn{1}{l}{9.66} & 2.62 \\ 
DTC & \multicolumn{1}{l}{16(20\%)} & 64 & \multicolumn{1}{l}{87.79} & \multicolumn{1}{l}{78.60} & \multicolumn{1}{l}{10.29} & 2.50 \\ 
CPS & \multicolumn{1}{l}{16(20\%)} & 64 & \multicolumn{1}{l}{88.72} & \multicolumn{1}{l}{80.10} & \multicolumn{1}{l}{7.49} & 1.91 \\ 
SASSNet & \multicolumn{1}{l}{16(20\%)} & 64 & \multicolumn{1}{l}{87.86} & \multicolumn{1}{l}{77.79} & \multicolumn{1}{l}{12.31} & 3.27 \\ 
MC-Net & \multicolumn{1}{l}{16(20\%)} & 64 & \multicolumn{1}{l}{89.18} & \multicolumn{1}{l}{79.94} & \multicolumn{1}{l}{6.52} & 1.66 \\ 
MT & \multicolumn{1}{l}{16(20\%)} & 64 & \multicolumn{1}{l}{89.01} & \multicolumn{1}{l}{81.21} & \multicolumn{1}{l}{6.08} & 1.96 \\ 
CEMT(Ours) & \multicolumn{1}{l}{16(20\%)} & 64 & \multicolumn{1}{l}{89.12} & \multicolumn{1}{l}{80.94} & \multicolumn{1}{l}{3.78} & 0.66 \\ 
UG-CEMT(Ours) & \multicolumn{1}{l}{16(20\%)} & 64 & \multicolumn{1}{l}{\textbf{89.73}} & \multicolumn{1}{l}{\textbf{81.63}} & \multicolumn{1}{l}{\textbf{2.20}} & \textbf{0.50} \\ \hline
\end{tabular}}

\end{table}
\textbf{Implementation details:}  With PyTorch library and A6000 NVIDIA GPU, we augmented the training data with random cropping to sub-volumes of size $112\times 112\times 80$. The training process utilizes the SAM optimizer with a base learning rate of 0.01, momentum of 0.9, weight decay of 0.0001, and a neighborhood size parameter $\rho=0.05$. We used a batch size of 4 with an equal distribution of labeled and unlabeled images. Our implementation includes a comprehensive set of parameters to manage consistency regularization effectively. We set $\lambda_{s}$ to 0.05 for balancing similarity loss and use a consistency ramp-up period of 40 epochs. The temperature for sharpening is set to 0.1. We use a memory bank with 256 embeddings per class, each embedding having a dimension of 64. Additionally, we filter 12,800 unlabeled embeddings to calculate. During training, we log the loss values and learning rate using TensorBoard and save the model checkpoints at regular intervals. The model is trained for a maximum of 6000 iterations, with the learning rate reduced by 0.1 every 2500 iterations.

\begin{table}[h!]
\centering
\caption{Comparison of the proposed UG-CEMT with other state-of-the-art methods on multi-site prostate dataset for 6000 iterations.}\label{sota-prostate}

\resizebox{10 cm}{!}{
\begin{tabular}{l llll}
\hline
\multicolumn{1}{c}{\multirow{2}{*}{Method}} & \multicolumn{2}{c}{(\%) of images used} & \multicolumn{2}{c}{Metrics} \\ \cline{2-5} 
\multicolumn{1}{c}{} & \multicolumn{1}{l}{labeled} & unlabeled & \multicolumn{1}{l}{Dice $\uparrow$} & Jaccard $\uparrow$ \\ \hline
V-VNet & \multicolumn{1}{l}{92(100\%)} & 0 & \multicolumn{1}{l}{78.76} & 66.52 \\ 
B-VNet & \multicolumn{1}{l}{92(100\%)} & 0 & \multicolumn{1}{l}{\textbf{80.76}} & \textbf{67.49} \\ 
V-VNet & \multicolumn{1}{l}{18(20\%)} & 0 & \multicolumn{1}{l}{64.63} & 54.87 \\ 
B-VNet & \multicolumn{1}{l}{18(20\%)} & 0 & \multicolumn{1}{l}{66.55} & 53.56 \\ \hline
UA-MT & \multicolumn{1}{l}{5(5\%)} & 87 & \multicolumn{1}{l}{62.16} & 52.58 \\ 
DTC & \multicolumn{1}{l}{5(5\%)} & 87 & \multicolumn{1}{l}{60.23} & 51.68 \\ 
CPS & \multicolumn{1}{l}{5(5\%)} & 87 & \multicolumn{1}{l}{61.58} & 53.36 \\ 
SASSNet & \multicolumn{1}{l}{5(5\%)} & 87 & \multicolumn{1}{l}{62.89} & 52.18 \\ 
MC-Net & \multicolumn{1}{l}{5(5\%)} & 87 & \multicolumn{1}{l}{61.34} & 51.23 \\ 
MT & \multicolumn{1}{l}{5(5\%)} & 87 & \multicolumn{1}{l}{61.57} & 51.78 \\ 
CEMT(Ours) & \multicolumn{1}{l}{5(5\%)} & 87 & \multicolumn{1}{l}{63.68} & 55.82 \\ 
UG-CEMT(Ours) & \multicolumn{1}{l}{5(5\%)} & 87 & \multicolumn{1}{l}{\textbf{65.68}} & \textbf{56.87} \\ \hline
UA-MT & \multicolumn{1}{l}{9(10\%)} & 83 & \multicolumn{1}{l}{65.67} & 57.13 \\ 
DTC & \multicolumn{1}{l}{9(10\%)} & 83 & \multicolumn{1}{l}{65.23} & 57.86 \\ 
CPS & \multicolumn{1}{l}{9(10\%)} & 83 & \multicolumn{1}{l}{64.17} & 56.15 \\ 
SASSNet & \multicolumn{1}{l}{9(10\%)} & 83 & \multicolumn{1}{l}{65.01} & 57.39 \\ 
MC-Net & \multicolumn{1}{l}{9(10\%)} & 83 & \multicolumn{1}{l}{64.23} & 56.69 \\ 
MT & \multicolumn{1}{l}{9(10\%)} & 83 & \multicolumn{1}{l}{66.04} & 56.52 \\
CEMT(Ours) & \multicolumn{1}{l}{9(10\%)} & 83 & \multicolumn{1}{l}{69.23} & 59.26 \\ 
UG-CEMT(Ours) & \multicolumn{1}{l}{9(10\%)} & 83 & \multicolumn{1}{l}{\textbf{70.36}} & \textbf{60.73} \\ \hline
UA-MT & \multicolumn{1}{l}{18(20\%)} & 74 & \multicolumn{1}{l}{68.43} & 59.68 \\ 
DTC & \multicolumn{1}{l}{18(20\%)} & 74 & \multicolumn{1}{l}{68.01} & 59.00 \\ 
CPS & \multicolumn{1}{l}{18(20\%)} & 74 & \multicolumn{1}{l}{67.56} & 58.86 \\ 
SASSNet & \multicolumn{1}{l}{18(20\%)} & 74 & \multicolumn{1}{l}{67.33} & 58.18 \\ 
MC-Net & \multicolumn{1}{l}{18(20\%)} & 74 & \multicolumn{1}{l}{67.89} & 58.67 \\ 
MT & \multicolumn{1}{l}{18(20\%)} & 74 & \multicolumn{1}{l}{67.36} & 57.24 \\ 
CEMT(Ours) & \multicolumn{1}{l}{18(20\%)} & 74 & \multicolumn{1}{l}{70.13} & 60.16 \\ 
UG-CEMT(Ours) & \multicolumn{1}{l}{18(20\%)} & 74 & \multicolumn{1}{l}{\textbf{72.02}} & \textbf{61.29} \\ \hline
\end{tabular}}
\end{table}
\textbf{Performance evaluation on LA dataset:} We compared UG-CEMT with the state-of-the-art SSL methods including UA-MT~\cite{yu2019uncertainty}, which leverages an uncertainty-guided approach; MC-Net~\cite{wu2021semi}, which utilizes mutual consistency learning with cycle pseudo-labels; DTC~\cite{luo2021semi}, which introduces multi-task consistency for medical image segmentation; SASSNet~\cite{li2020shape}, which incorporates geometric constraints into the network; CPS~\cite{chen2021semi} which uses cross pseudo supervision approach. Additionally, we implemented MT-based UA-MT for a more comprehensive comparison~\cite{tarvainen2017mean}.
The results are presented in Table~\ref{sota-la}. We evaluated our model on different (\%) of labeled data, including 5\%, 10\%, and 20\%. We also reported the performance of V-VNet (Vanilla-VNet) and B-VNet (Bayesian-VNet) by increasing the dropout at 100\% and 20\% labeled data to serve as reference upper bounds and baselines. As shown in Table~\ref{sota-la}, all SSL methods benefit from incorporating unlabeled data. UA-MT outperformed the MT, indicating that the uncertainty map can enhance the student model's performance. Among the comparison methods, MC-Net demonstrated the best results in terms of Dice and Jaccard with stable performance for 20\% labeled data. Overall, CPS excelled in 95HD and ASD metrics compared to all existing methods, and SSANet performed well, suggesting that incorporating shape priors can enhance edge segmentation. The proposed UG-CEMT framework outperformed all state-of-the-art methods across all metrics and also substantially improved results on edge-sensitive metrics such as 95HD and ASD were obtained. Compared to SASSNet, UG-CEMT reduced 95HD from 12.31 to 2.30 mm and ASD from 3.27 to 0.50 mm for 20\% labeled data, and a similar range of decrement was there for 10\% and 5\% labeled data, indicating a more stable and accurate performance.
\begin{figure*}[h!]
    \centering
    \includegraphics[width=0.9\linewidth]{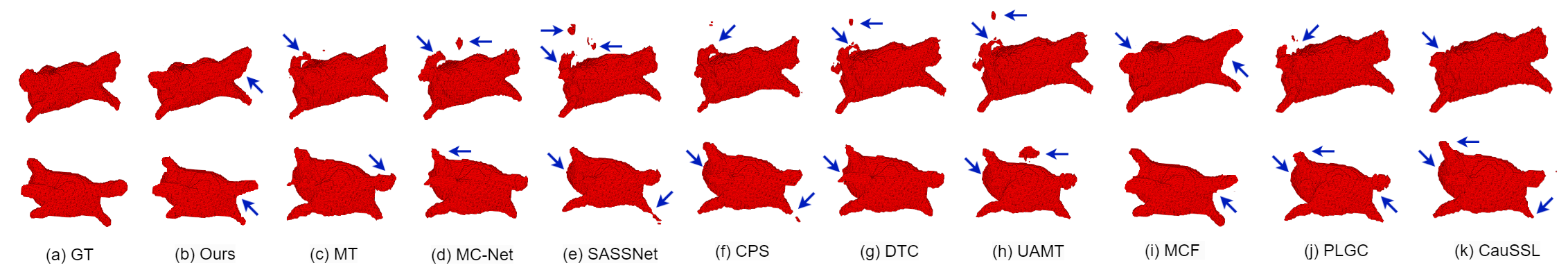}
    \caption{Visualization of 3D segmentation outcomes of various SSL methods for $20\%$ labeled data on LA dataset.} 
    \label{fig:LA}
\end{figure*}

\begin{figure*}[h!]
    \centering
    \includegraphics[width=0.9\linewidth]{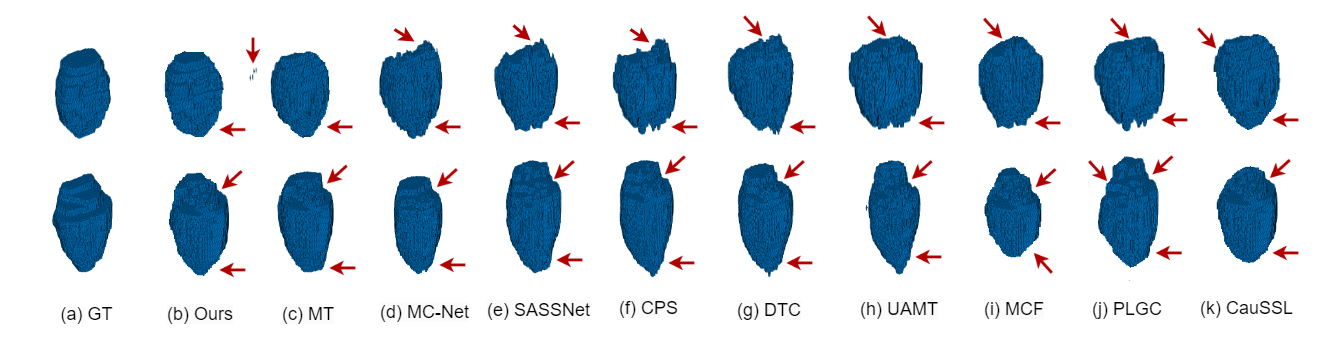}
    \caption{Visualization of 3D segmentation outcomes of various SSL methods for $20\%$ labeled data on multi-site prostate dataset.}
    \label{fig:prost}
\end{figure*}

\textbf{Performance evaluation on multi-site prostate dataset:}
Prostate segmentation is challenging due to small and variable organ shape, scanner, and other physiological variations as well as the domain generalization issues that arise from multi-source datasets.
As illustrated in Table~\ref{sota-prostate}, we provided the results at different (\%) of labeled images. Like LA segmentation, B-VNet outperforms V-VNet in a fully supervised manner. UA-MT outperforms all the other existing methods, such as dice and Jaccard. Notably, the performance of these existing approaches on the two datasets is inconsistent; it shows which method performs worse in LA segmentation and describes an advantage on the multi-site prostate datasets, such as UA-MT, DTC, and SSANet. The gap between LA and prostate segmentation performance is more because of the complexity of multi-source data. Our proposed method consistently showed better results, on the other hand.

\textbf{Qualitative analysis:} Figures \ref{fig:LA} and \ref{fig:prost} display some of the LA and prostate segmentation results. As shown, CPS and SASSNet on LA segmentation, and MC-Net and CPS on prostate segmentation, tend to under-segment specific regions, likely due to restricted generalization abilities. MCF-Net~\cite{wang2023mcf}, PSGC~\cite{basak2023pseudo}, and CauSSL~\cite{miao2023caussl} show comparable performance with our model. In contrast, our UG-CEMT framework produces more precise outcomes, capturing finer segmentation features with a more effective training strategy. 
Our model generates relatively high-confidence predictions from the UGM images. Initially, UG-CEMT produces high uncertainty guided maps (For additional visual results, please refer to the \textbf{\textit{supplementary material}} Figure 1), but progressively enhances its confidence in the input images during training. These results demonstrate that UGM can enable SSL approaches to produce high-confidence predictions, ensuring more effective co-training with UG-CEMT. Additionally, we conducted additional experiments using a pancreas CT dataset~\cite{roth2015deeporgan} to further assess the generalizability of UG-CEMT across different imaging modalities. The qualitative and quantitative results of these experiments, along with detailed dataset descriptions and analysis, can be found in the \textbf{\textit{supplementary material.}}

\begin{figure*}[ht]
    \centering
    \includegraphics[width=0.8\linewidth]{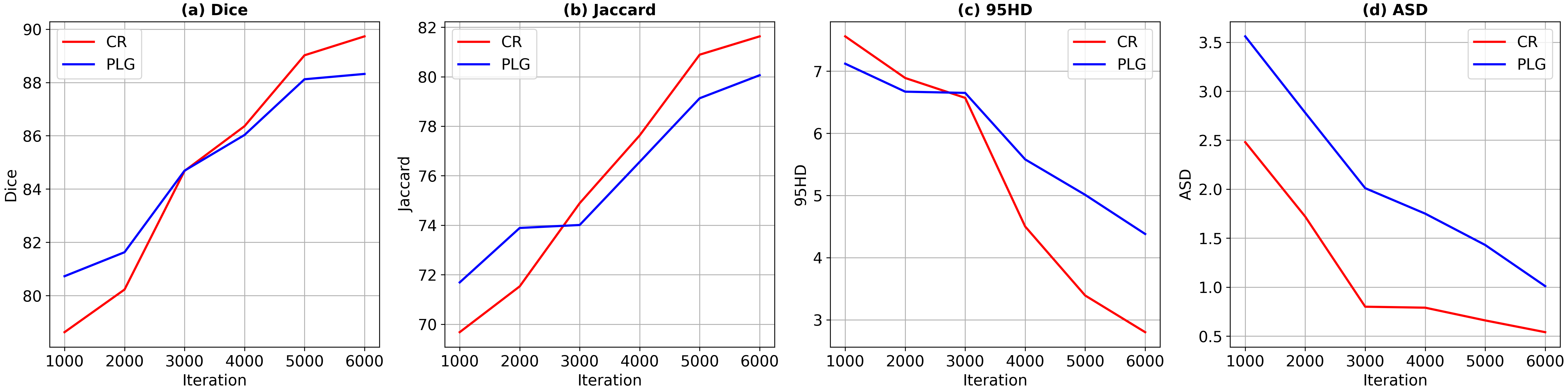}
    \caption{Performance comparison between pseudo label generation (PLG) and consistency regularization (CR) label training}
    \label{fig:CR}
\end{figure*}
 
\begin{table}[ht]
\centering
\caption{Performance comparison when different regularization techniques used during training on LA dataset.\label{regularizers}}
\resizebox{7 cm}{!}{
\begin{tabular}{c c c c c}
\hline
Regulizer & Dice $\uparrow$  & Jaccard $\uparrow$ & 95HD $\downarrow$ & ASD $\downarrow$\\  \hline
L1 & 88.02 & 79.96 & 6.08 & 2.16 \\
L2 & 88.34 & 79.82 & 5.76 & 3.68 \\ 
SAM & \textbf{89.73} & \textbf{81.63} & \textbf{2.8} & \textbf{0.5} \\ \hline
\end{tabular}}
\end{table}
\begin{table}[ht]
\centering
\caption{Ablation study for combining different components on LA dataset. ST: student-teacher, CA: cross-attention, EWA: exponential weighted average, U: uncertainty guided map. \label{ablation}}
\resizebox{8 cm}{!}{
\begin{tabular}{clllllll}
\hline
\multicolumn{1}{l}{labeled data} & Study & ST & CA & EWA & U & Dice $\uparrow$ & 95HD $\downarrow$ \\ \hline
5\% & Baseline & $\checkmark$ &  &  &  & 78.16 & 11.89 \\ 
 & MT & $\checkmark$ &  & $\checkmark$ &  & 83.72 & 6.37 \\ 
 & CEMT & $\checkmark$ & $\checkmark$ &  $\checkmark$&  & 85.23 & 5.12 \\ 
 & UG-CEMT & $\checkmark$ & $\checkmark$ & $\checkmark$ & $\checkmark$ & \textbf{85.89} & \textbf{3.39} \\ \hline
10\% & Baseline &$\checkmark$  &  &  &  & 84.13 & 9.15 \\ 
 & MT & $\checkmark$ &  & $\checkmark$ &  & 86.28 & 5.78 \\ 
 & CEMT & $\checkmark$ & $\checkmark$ &  $\checkmark$&  & 87.03 & 3.39 \\ 
 & UG-CEMT & $\checkmark$ & $\checkmark$ & $\checkmark$ & $\checkmark$ & \textbf{88.16} & \textbf{3.08} \\ \hline
20\% & Baseline & $\checkmark$ &  &  &  & 85.17 & 7.18 \\ 
 & MT & $\checkmark$ &  & $\checkmark$ &  & 87.89 & 4.68 \\ 
 & CEMT & $\checkmark$ & $\checkmark$ & $\checkmark$ &  & 89.02 & 3.78 \\ 
 & UG-CEMT & $\checkmark$ & $\checkmark$ & $\checkmark$ & $\checkmark$ & \textbf{89.73} & \textbf{2.20} \\ \hline
\end{tabular}}
\end{table}
\textbf{Computational cost analysis:} The proposed UG-CEMT model consists of 9.66M parameters and requires 47.1G MACs (FLOPs) for a single forward pass. Despite the complexity introduced by the cross-attention mechanism, the model achieves significant performance improvements while remaining computationally feasible. The total training time of $\approx$ 1 hour 50 minutes for 6000 iterations, demonstrates the model's efficiency and suitability for real-world deployment. A detailed analysis is provided in the \textbf{\textit{supplementary material.}}
\subsection{Ablation study}
We conducted ablation experiments to demonstrate the effectiveness of each component in our UG-CEMT framework including regularization methods.

\textbf{Effect of regularization techniques:}  We examined the impact of different regularization techniques on the performance of our UG-CEMT framework on LA dataset for 20\% labeled images for 6000 iterations. As shown in Table~\ref{regularizers}, we compared L1 and L2 regularization with SAM. Using L1 regularization, the model achieved a Dice score of 88.02\% and a 95HD of 6.08 mm. L2 regularization slightly improved the performance, with a Dice score of 88.34\% and a 95HD of 5.76 mm. However, SAM significantly outperformed both L1 and L2 methods, achieving the highest Dice score of 89.73\% and the lowest 95HD of 2.8 mm. These results demonstrate that SAM effectively enhances model generalization and performance, making it a superior regularization technique for our framework.

\textbf{Effects of different components:} As shown in Table~\ref{ablation}, we conducted ablation experiments to demonstrate the effectiveness of each component in our UG-CEMT framework for 6000 iterations. The baseline ST setup achieved a Dice score of 78.16\% and a 95HD of 11.59 mm with 5\% labeled data. Adding EWA improved performance to a Dice score of 83.72\% and a 95HD of 6.37 mm. 
Incorporating CA further enhanced performance, achieving a Dice score of 85.23\% and a 95HD of 5.12 mm. The complete UG-CEMT framework, integrating all components, achieved the highest Dice score of 85.89\% and the lowest 95HD of 3.39 mm with 5\% labeled data. This demonstrates that each component contributes to overall performance improvement, with the full model yielding the best results (For qualitative results, please refer to the \textbf{\textit{supplementary material}} Figure 2).

\textbf{CR and PLG:} To understand dynamic performance changes during training, we compared consistency regularization (CR) with pseudo label generation (PLG) in our UG-CEMT framework. Figure~\ref{fig:CR} shows performance metrics over iterations for Dice, Jaccard, 95HD, and ASD on the LA dataset with 20\% labeled images. Initially, PLG outperformed CR within the first 2,000 iterations. However, beyond 2,000 iterations, CR surpassed PLG, indicating its effectiveness in maintaining high performance as training progresses. After 5,000 iterations, the gap widened, with CR demonstrating superior performance across all metrics, highlighting its robustness in leveraging unlabeled data for consistent performance improvement.

\section{Discussion and Concluding Remarks}
In this work, we presented UG-CEMT, a novel framework for semi-supervised medical image segmentation that leverages uncertainty-guided cross-attention ensemble mean teacher learning. Our experiments on challenging 3D left atrium and multi-site prostate MRI datasets demonstrate that UG-CEMT outperforms state-of-the-art SSL methods across various metrics and labeled data ratios. The superior performance of UG-CEMT can be attributed to several key factors including (1) enhanced feature alignment, (2) uncertainty-guided learning, and (3) improved generalization. Our ablation studies reveal the importance of each component in the UG-CEMT framework. The combination of cross-attention, exponential weighted average, and uncertainty-guided maps consistently yields the best performance across different labeled data ratios.  The comparison between consistency regularization (CR) and pseudo-label generation (PLG) highlights the long-term stability and effectiveness of CR in leveraging unlabeled data. This finding suggests that carefully designed consistency constraints can be more beneficial than relying solely on generated pseudo-labels, especially as training progresses. The UG-CEMT makes it particularly suitable for clinical applications, where obtaining large labeled datasets is challenging due to the need for expert annotations. By leveraging a combination of labeled and unlabeled data, UG-CEMT reduces annotation costs while maintaining high performance, which is critical in real-world medical imaging workflows. While UG-CEMT shows promising results, there are limitations and areas for future work. \textit{Computational complexity:} The cross-attention mechanism and uncertainty estimation increase computational overhead. Future work could explore more efficient implementations or lightweight alternatives. \textit{Generalization to other tasks:} While we focused on left atrium and prostate segmentation, further studies could investigate the applicability of UG-CEMT to a broader range of medical imaging tasks and modalities. \textit{Uncertainty calibration:} Although our method leverages uncertainty estimates, further research into calibrating these uncertainties could potentially lead to even more reliable predictions.\\
\textbf{Acknowledgments}
This study is supported by NIH grants: R01-CA246704, R01 CA240639, R01-HL171376, and U01-CA268808.

\bibliographystyle{unsrtnat}
\bibliography{references}  






\end{document}